\documentclass[letterpaper, 10 pt, conference]{ieeeconf}  % Comment this line out if you need a4paper

\IEEEoverridecommandlockouts                              % This command is only needed if 
\usepackage{graphics} % for pdf, bitmapped graphics files
\usepackage{epsfig} % for postscript graphics files
\usepackage{mathptmx} % assumes new font selection scheme installed
\usepackage{times} % assumes new font selection scheme installed
\usepackage{amsmath} % assumes amsmath package installed
\usepackage{amssymb}  % assumes amsmath package installed
\usepackage{caption}
\usepackage{hyperref} %JE
\usepackage{xcolor} %JE
\usepackage{multirow} %JE
\usepackage{colortbl} %JE
\usepackage{siunitx}
\usepackage{booktabs}
\usepackage{tabularx}
\newcommand{\PIER}[1]{\textcolor{blue} {#1} } %PIER
\usepackage{bm}
\DeclareMathOperator*{\argmin}{arg\,min} %JED

\title{\LARGE \bf CT-ICP: Real-time Elastic LiDAR Odometry with Loop Closure}
\author{Pierre Dellenbach$^{1,2}$, Jean-Emmanuel Deschaud$^{1}$, Bastien Jacquet$^{2}$, and François Goulette$^{1}$
\thanks{$^{1}$ MINES ParisTech, PSL University, Centre for Robotics, 75006 Paris, France
\{firstname.surname@mines-paristech.fr\}}
\thanks{$^{2}$ Kitware, Computer Vision Team, 69100 Villeurbanne, France, \{firstname.surname@kitware.com\}}
}

\newcommand{\NCD}{NCD}
\newcommand{\NCLT}{NCLT}
\newcommand{\KITTI}{KITTI}
\newcommand{\newKITTI}{KITTI-360}
\newcommand{\KITTICarla}{KITTI-CARLA}

\newcommand{\ParisLuco}{ParisLuco}

\definecolor{strong_green}{RGB}{0,136,6}
\definecolor{light_green}{RGB}{0,200,6}
\newcommand\col[1]{\textcolor{strong_green}{\textbf{#1}}}
\definecolor{strong_red}{RGB}{136,0,6}

\newcommand\customfigcaption[1]{\small\textit{ #1}}
\newcommand\customtabcaption[1]{\small\textit{ #1}}
\newcommand\tabspace{-20pt}
\newcommand\figspace{-15pt}
\newcommand\greenfootnote[1]{\textcolor{light_green}{\footnote{\textcolor{light_green}{\url{#1}}}}}

\begin{document}

\maketitle
\thispagestyle{empty}
\pagestyle{empty}

%%%%%%%%%%%%%%%%%%%%%%%%%%%%%%%%%%%%%%%%%%%%%%%%%%%%%%%%%%%%%%%%%%%%%%%%%%%%%%%%
\begin{abstract}

Multi-beam LiDAR sensors are increasingly used in robotics, particularly with autonomous cars for localization and perception tasks, both relying on the ability to build a precise map of the environment. 
For this, we propose a new real-time LiDAR-only odometry method called CT-ICP (for Continuous-Time ICP), completed into a full SLAM with a novel loop detection procedure. 
The core of this method, is the introduction of the combined continuity in the scan matching, and discontinuity between scans. 
It allows both the elastic distortion of the scan during the registration for increased precision, and the increased robustness to high frequency motions from the discontinuity.

% The principle of CT-ICP is to use an elastic formulation of the trajectory, with a continuity of poses intra-scan and discontinuity between scans, to be more robust to high frequencies in the movements of the sensor. 
We build a complete SLAM on top of this odometry, using a fast pure LiDAR loop detection based on elevation image 2D matching, providing a pose graph with loop constraints. 
To show the robustness of the method, we tested it on seven datasets: KITTI, KITTI-raw, KITTI-360, KITTI-CARLA, ParisLuco, Newer College, and NCLT in driving and high-frequency motion scenarios.
Both the CT-ICP odometry and the loop detection are made available online. 
CT-ICP is currently first, among those giving access to a public code, on the KITTI odometry leaderboard, with an average Relative Translation Error (RTE) of 0.59\% and an average time per scan of 60$ms$ on a CPU with a single thread.

\end{abstract}

%%%%%%%%%%%%%%%%%%%%%%%%%%%%%%%%%%%%%%%%%%%%%%%%%%%%%%%%%%%%%%%%%%%%%%%%%%%%%%%%
\section{INTRODUCTION}
% main author JE

%\PIER{For cameras the rolling shutter problem can be resolved in ardware using global shutter cameras which acquire all pixels intensities synchronously, however for LiDAR sensors it is deeply embedded in the technology. Most LiDARs capture spatial information continuously, with several laser fibers mounted on moving parts to extend the geometry scanned by the sensor. }

%\JE{Unlike conventional cameras which acquire an entire image at a certain frequency (\PIER{Rolling shutter is still a problem for cameras without Global shutter} i.e., all pixels in the image are acquired at the same time), LiDAR sensors acquire data continuously. Indeed, most LiDARs are based on the principle of several laser fibers which shoot continuously on a rotating mechanism.}

Many LiDARs are based on the principle of several laser fibers that fire off continuously out of a rotating unit.
We call a ``scan" the limited time-span aggregation of points covering enough field of view (360 degrees for LiDARs used in autonomous vehicles like in~\cite{geiger2012kitti, nuscenes2019}).

To take continuous acquisition into account, LiDAR odometry methods like in~\cite{wang2021floam} or~\cite{pan2021mulls} distort the current scan with a constant velocity motion assumption. However, this assumption does not take into account large direction or velocity changes.

On an other hand, recent works like~\cite{park2018elastic, quenzel2021ct} propose real-time LiDAR odometries with continuous-time trajectories. A pose can be calculated for each LiDAR point according to its timestamp from control poses (direct poses or using splines). 
However, \cite{park2018elastic} requires an IMU to integrate high-frequency motions, while the continuity constraints in \cite{quenzel2021ct} simply smoothes out these motions, we believe, at the cost of precision.

We propose a new elastic formulation of the trajectory with a continuity of poses intra-scan and discontinuity between adjacent scans. 
In practice, this is defined by solving an elastic scan-to-map registration, parametrized by two poses per scan (for the beginning and end of the scan) with a proximity constraint between the end pose of the previous scan and the beginning pose of the current scan.

%trim <left> <lower> <right> <upper>
\begin{figure}[t!]
    \centering
    \includegraphics[trim=0 0 200 80, clip=true, width=\linewidth]{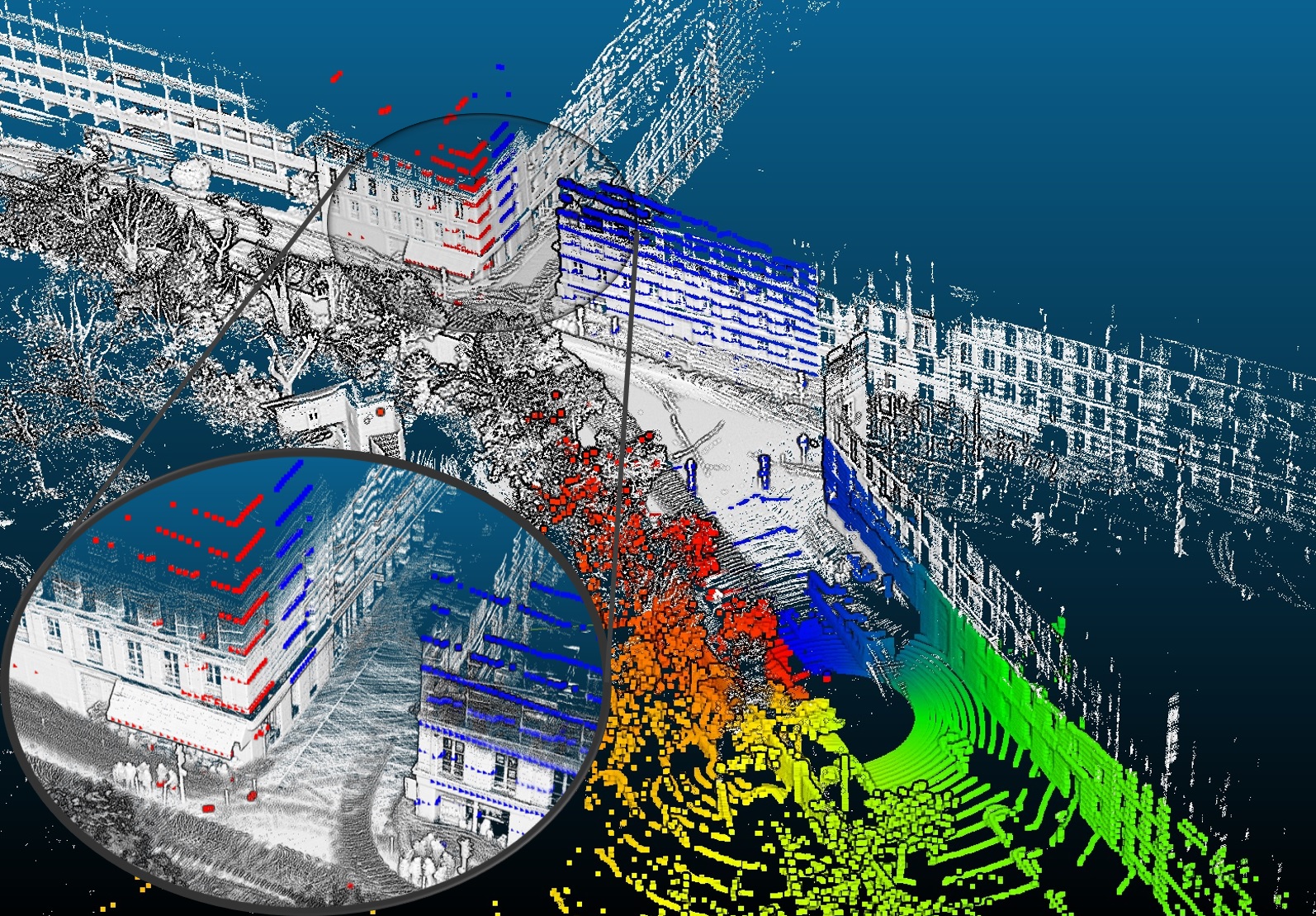}
    \includegraphics[trim=0 42 0 0, clip=true, width=\linewidth]{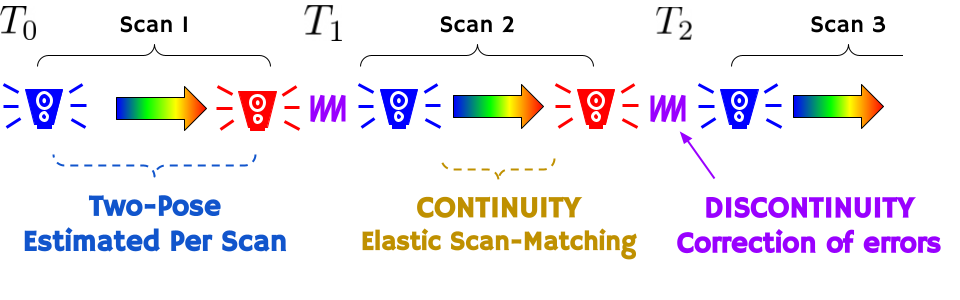}
    \caption{\customfigcaption{Top, in color, one LiDAR scan; the color depends on the timestamp of each point (from the oldest in blue to newest in red). The scan is deformed elastically to align with the map (white points) by the joint optimization of two poses at the start and end of the scan and interpolation according to the timestamp, hence creating a continuous-time scan-to-map odometry. Below, the formulation of our trajectory with a continuity of poses intra-scan and discontinuity between scans.}}
    
    \label{fig:fig_ct_icp}
    \vspace{\figspace}
\end{figure}

Fig.~\ref{fig:fig_ct_icp} shows the registration of a scan (points are in color) to the map (points are in white). The color ranges from blue to red, representing the relative timestamp ($\alpha_i$) for each point used by our CT-ICP method. Our odometry elastically adjusts the new scan to the building.

As our main contribution, we propose:
\begin{itemize}
\item A new elastic LIDAR odometry based on the continuity of poses intra-scan and discontinuity between scans.
\end{itemize}

We also present as secondary contributions:
\begin{itemize}
    \item A local map based on a dense point cloud stored in a sparse voxel structure to obtain real-time processing speed.
    \item A large campaign of experiments on 7 datasets in driving and high-frequency motion scenarios, all reproducible with public and permissive open-source code \greenfootnote{https://github.com/jedeschaud/ct_icp}.
    \item A fast method of loop detection integrated with a pose graph back-end to build a complete SLAM, integrated into \textbf{\texttt{pyLiDAR-SLAM}} \greenfootnote{https://github.com/Kitware/pyLiDAR-SLAM}.
\end{itemize}

%%%%%%%%%%%%%%%%%%%%%%%%%%%%%%%%%%%%%%%%%%%%%%%%%%%%%%%%%%%%%%%%%%%%%%%%%%%%%%%%
\section{RELATED WORK}
% main author JE

Numerous LiDAR odometry methods are based on the Iterative Closest Point (ICP) method~\cite{besl1992icp} and its more efficient point-to-plane variant~\cite{rusinkiewicz2001variantsicp, pomerleau2015review}. KinectFusion~\cite{newcombe2011kinectfusion} for RGB-D sensors has shown great progress in moving from frame-to-frame to frame-to-model registration, likewise, LiDAR odometry methods use scan-to-map registration.

SuMa~\cite{behley2018suma} and SuMa++~\cite{chen2019sumaplusplus} represent a scan as an image (range image) and the map in the form of a set of surfels. The registration aligns the current scan with a rendered image obtained by projecting the map of surfels on the GPU. LOAM~\cite{zhang2017loam} detects keypoints of different classes (edges, planes) in the range image and registers the detected keypoints into a voxel grid, but the neighborhood search uses separate kd-trees for each class of keypoint. The keypoint map is sufficiently sparse for the search to be carried out in real time. LeGO-LOAM~\cite{shan2018legoloam} improves on the previous method by separating the keypoints from the ground. F-LOAM~\cite{wang2021floam} has optimized the registration to be faster and runs at more than 20\,Hz. More recently, MULLS~\cite{pan2021mulls} has improved this approach by detecting the keypoints in the scan in 3D with many different types of keypoints (ground, facade, roof, pillar, beam, and vertex). Differently, IMLS-SLAM~\cite{deschaud2018imlsslam} represents the map in the form of a dense point cloud,~\cite{kuhner2020tsdf} in the form of a TSDF, and PUMA~\cite{vizzo2021puma} with a mesh, but these last three methods do not work in real time.

To account for sensor movement during scanning, most previous methods like~\cite{wang2021floam, pan2021mulls, deschaud2018imlsslam} first distort the current scan with a constant velocity motion model from previous poses. 
Then, the scan distortion is kept fixed during ICP iterations. 
Although this approach works well in most driving scenarios, it is not robust enough to sudden changes of orientation or fast acceleration from one scan to another.

%All previous methods formulate the trajectory in the form of a pose $(R, t)$ to align the current scan with the map.  LOAM~\cite{zhang2017loam} and LeGO-LOAM~\cite{shan2018legoloam} perform a first scan-to-scan registration to estimate the correction before doing a scan-to-map. Other methods (Velodyne SLAM~\cite{moosmann2011}, F-LOAM~\cite{wang2021floam}, MULLS~\cite{pan2021mulls}) 
%~\cite{ceriani2015moving} represents the trajectory in a linear continuous-time way and the map in the form of keypoints stored in a voxel hash map.

Another approach takes into account the motion during scanning by defining a continuous-time trajectory with control poses (by linear interpolation or B-splines). 
CT-SLAM~\cite{bosse2009continuous} defines the trajectory by using multiple poses per scan;~\cite{hatem2014continous} also uses a continuous-time trajectory with six poses per scan (using B-splines basis to represent the trajectory). 
However, both methods are not in real time. 
More recently, Elastic LiDAR Fusion in~\cite{park2018elastic} and MARS LiDAR odometry in~\cite{quenzel2021ct} proposed a continuous-time formulation of the trajectory that operates in real time.~\cite{park2018elastic} uses a linear interpolation of poses with a map of sparse surfels for odometry and dense 2D disk surfels for mapping, while~\cite{quenzel2021ct} leverages B-splines for the continous-time trajectory with a multi-resolution surfel map to get real-time speed. 
These methods tend to smooth the trajectory, but in many real-life acquisitions, the motion of the sensor can be shaky (notably due to terrain irregularities) and produce high-frequency movements (with respect to the control point frequencies), which are not taken into account by these methods. 

By contrast, CT-ICP defines a continous-time trajectory during the scan and discontinuous between scans. During a scan, the trajectory is parameterized by two poses for the beginning and end of the scan. However, conversely to methods like~\cite{quenzel2021ct}, the final trajectory of CT-ICP is discontinuous, the pose at the beginning of a scan is not equal to the pose at the end of the previous scan. We claim that this approach compensates for the motion irregularities that cannot be compensated for by interpolation.

%The  before the registration step by ICP and make a new estimate of the distortion with the final pose. To correct the distortion of the current scan, some methods use LiDAR-IMU coupling like LIO-SAM~\cite{shan2020liosam} or LiLi-OM~\cite{li2021liliom}.

%For IMLS-SLAM~\cite{deschaud2018imlsslam}, the keypoints of the current scan are selected by geometric criteria of planarity and observability. The registration of the current scan with the dense point cloud map from previous scans uses a kd-tree type search structure. The results produced are more precise than the range image or sparse keypoints approaches but the method is not real time because of the search by kd-tree in the dense point cloud map.

%Recent Deep Learning based LiDAR odometry methods using a PoseNet like DeepLO~\cite{cho2020deeplo} have shown promising results but are not yet at the accuracy level of classical methods as shown by~\cite{dellenbach2021}. Hybrid methods like LO-Net~\cite{li2019lonet} (using PoseNet as initialization and then a classical scan-to-model) or DMLO~\cite{li2020dmlo} (U-Net netwok for features extraction then classical matching) have shown interesting results but are still limited for the generalization on new datasets.

As precise as it is, LiDAR odometry still accumulates errors in an open environment, which leads to drift in trajectories. A loop closure procedure can correct the trajectories globally, but loop detection is still an open problem for LiDAR SLAM. Currently, most SLAM solutions rely principally on registration methods to directly close loops \cite{mendes2016icp, behley2018suma, zhang2017loam}, which only works for small trajectories and low drift. 
Different place recognition methods have been proposed, operating on individual scans \cite{scan_context, bvmatch, overlap_net, one_day_one_year}, which is sensitive to environment changes and are more adapted for driving scenarios.~\cite{pan2021mulls} uses a global registration procedure \cite{teaser} before ICP-based refinement, but each alignment attempt is very costly, thus restricting the chances to discover more loops. 
Recently, deep learning methods have been proposed \cite{overlap_net, one_day_one_year}, but are not adapted to new environments due to the training requirements. 
By contrast, we propose a new loop closure procedure that operates on aggregated point clouds projected onto an elevation image. 
This procedure requires the motion of the sensor to be mostly 2D and to estimate the gravity vector; thus, it can be integrated into any LiDAR odometry for which these conditions are met. 
Closest to our work is \cite{bvmatch}, which constructs elevation images but operates on scans, whereas our procedure runs on local maps and, therefore, is not required to test each scan against the previous observed locations, making it more efficient for online SLAM scenarios.

%%%%%%%%%%%%%%%%%%%%%%%%%%%%%%%%%%%%%%%%%%%%%%%%%%%%%%%%%%%%%%%%%%%%%%%%%%%%%%%%
\section{CT-ICP ODOMETRY} \label{ct_icp_method}
% main author JE

%\PIER{TODO: Search for article having been evaluated on KITTI-360, NCLT, KITTI (recents)} 
\begin{figure}
%\includegraphics[width=\linewidth]{images/aggregated_clouds.png}
%trim <left> <lower> <right> <upper>
\includegraphics[trim=0 0 0 0, clip=true, width=0.24\textwidth]{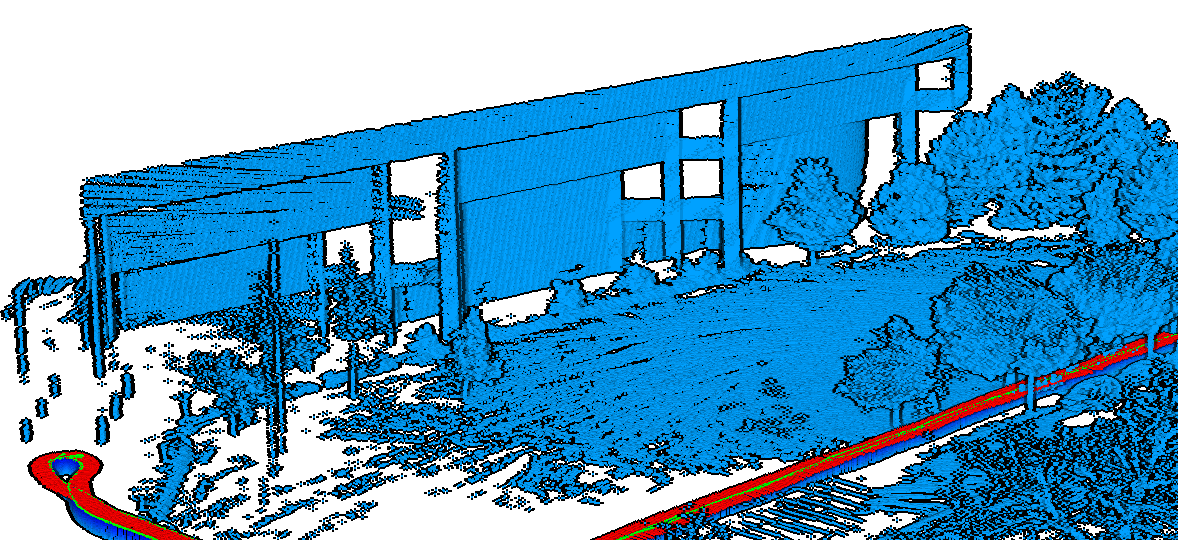}
\includegraphics[trim=0 0 50 0, clip=true, width=0.24\textwidth]{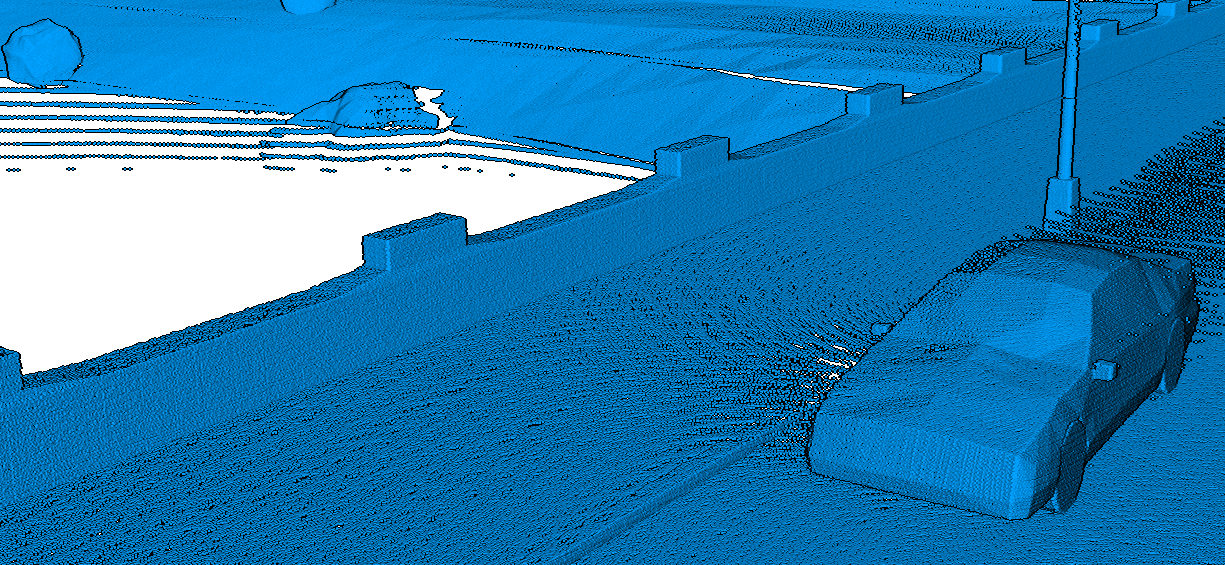}
\includegraphics[trim=0 0 0 0, clip=true, width=0.24\textwidth]{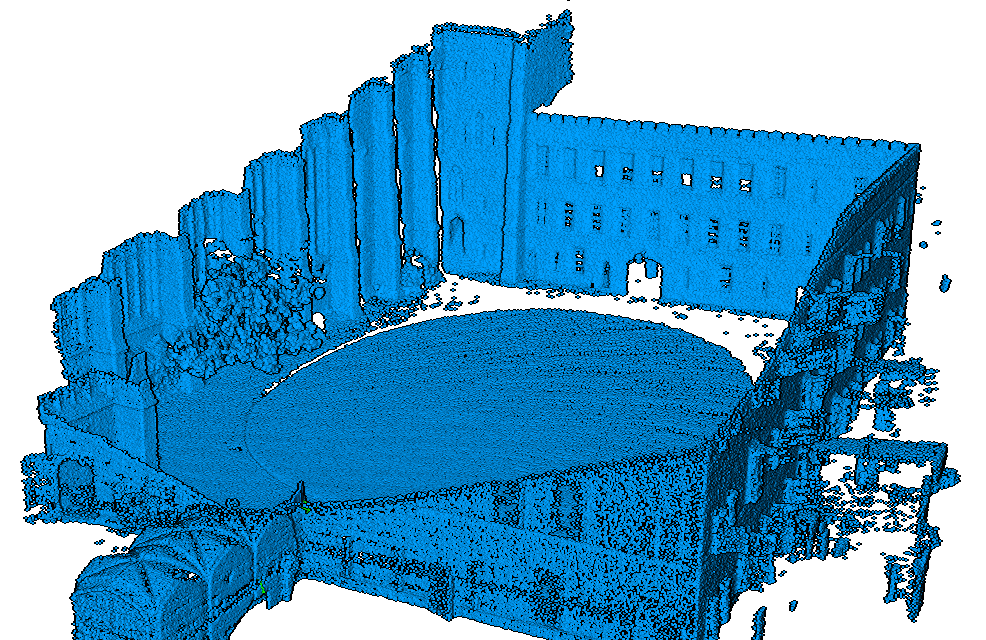}
\includegraphics[trim=0 0 400 200, clip=true, width=0.24\textwidth]{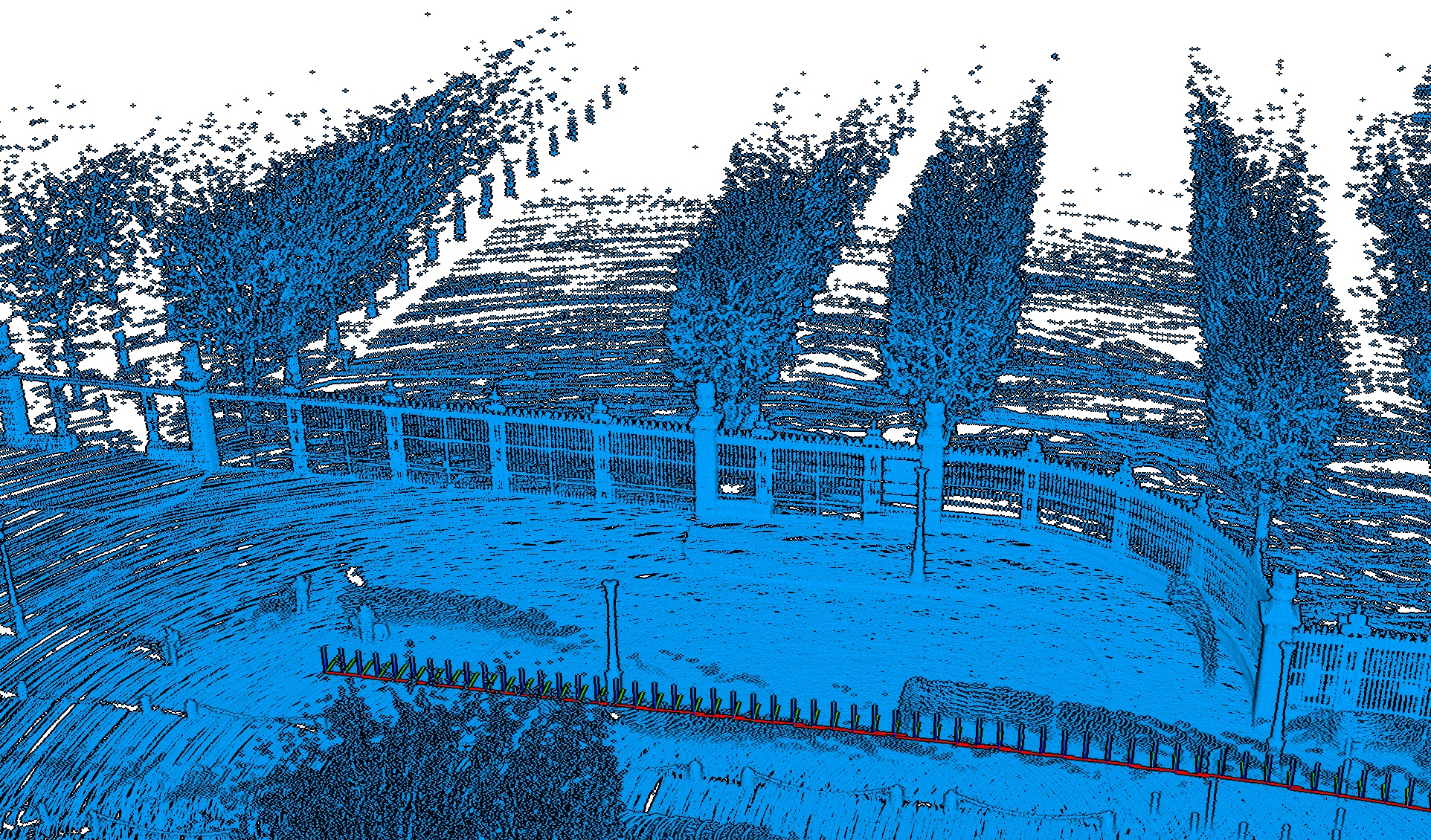}
\caption{\customfigcaption{Aggregated point clouds for NCLT dataset (top left), KITTI-CARLA (top-right), Newer College Dataset (bottom left), and ParisLuco (bottom right) show the quality of the maps obtained with CT-ICP.}}
\label{nclt}
\vspace{-10pt}
\end{figure}

\subsection{Odometry formulation}
\begin{figure*}[ht]
\centering
\includegraphics[width=0.19\linewidth]{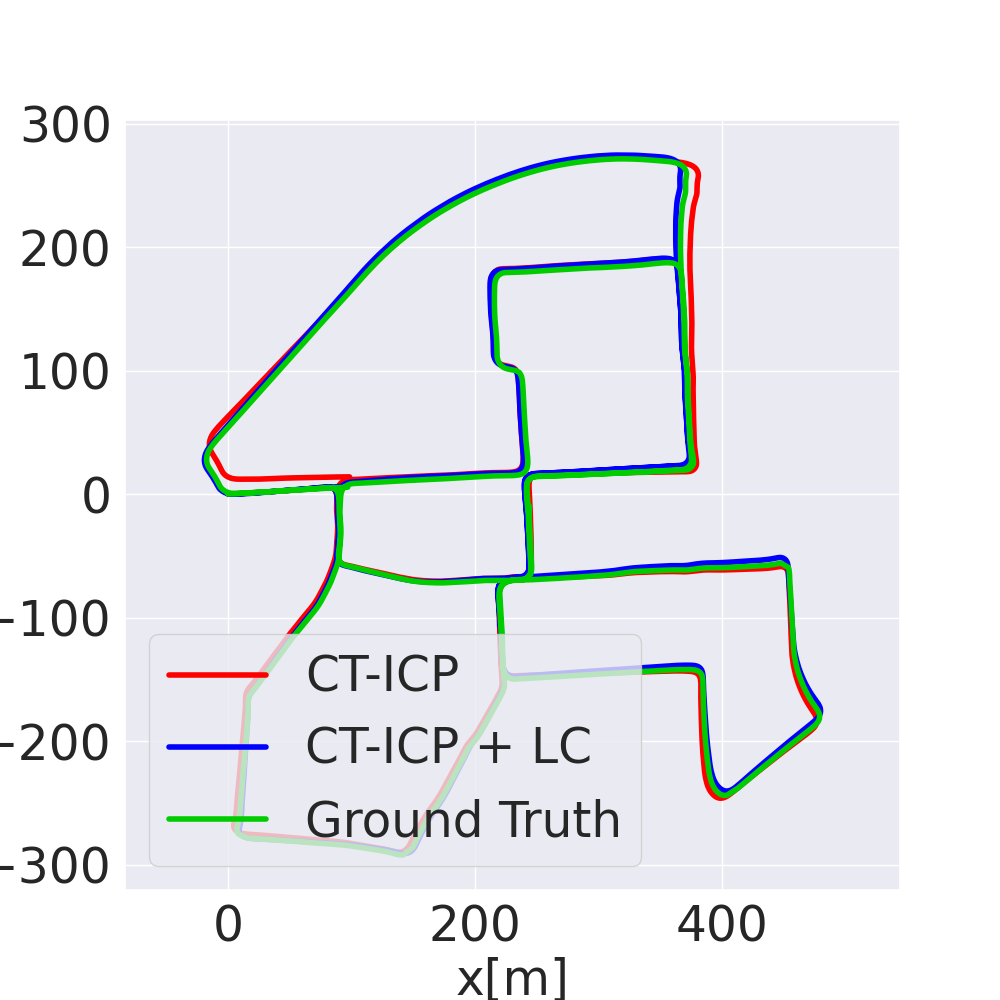} 
\includegraphics[width=0.19\linewidth]{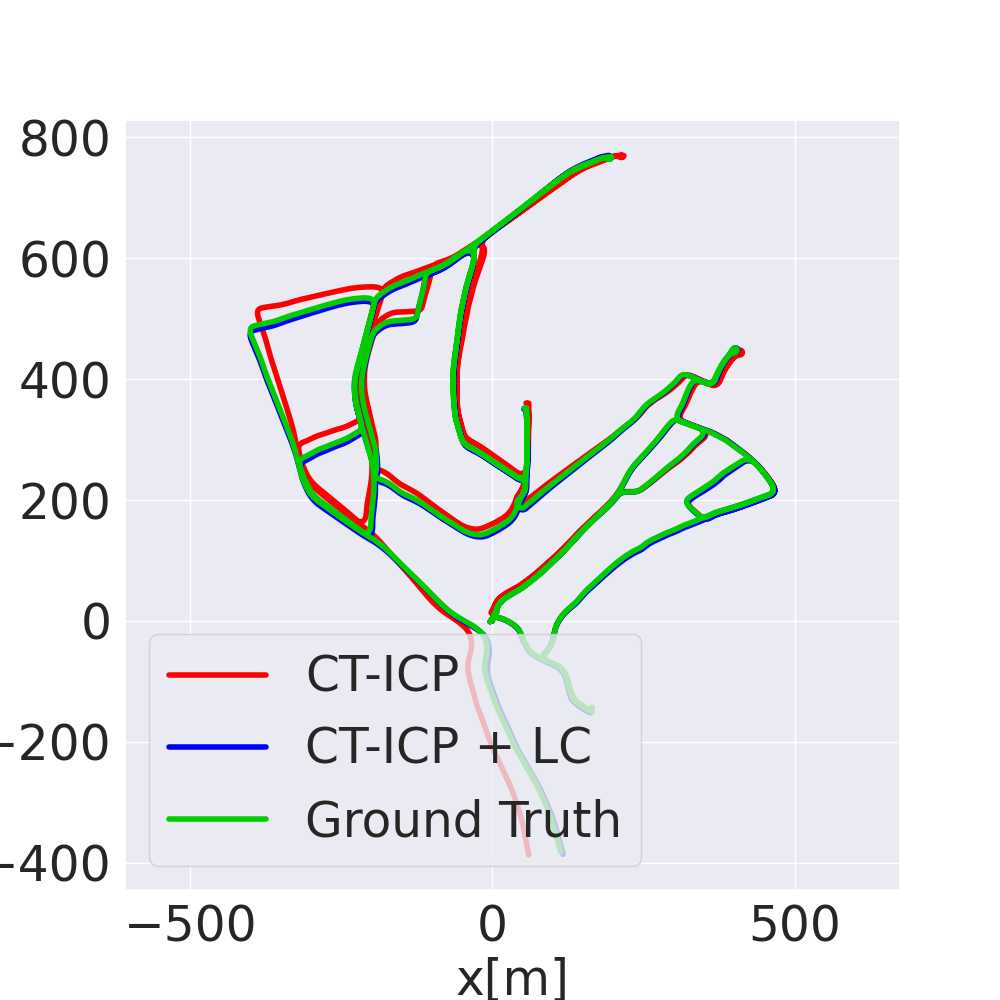} 
\includegraphics[width=0.19\linewidth]{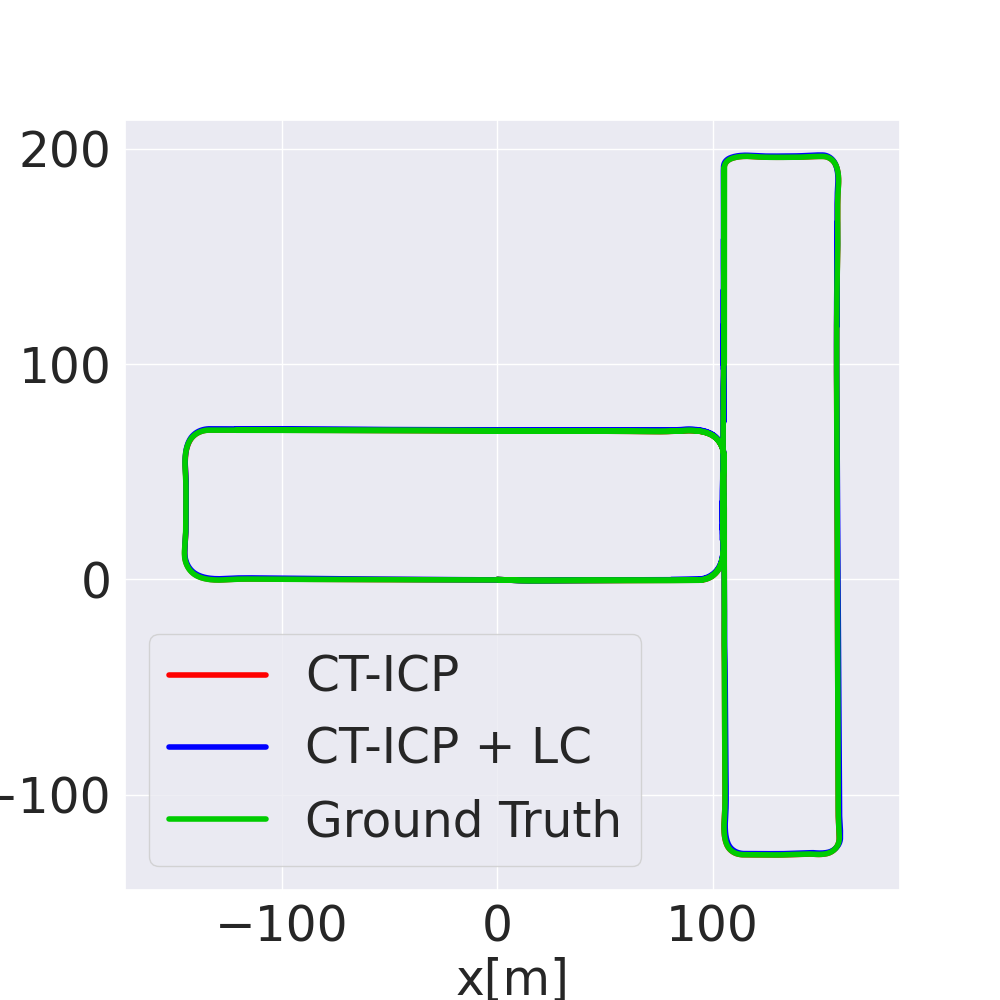}
\includegraphics[width=0.19\linewidth]{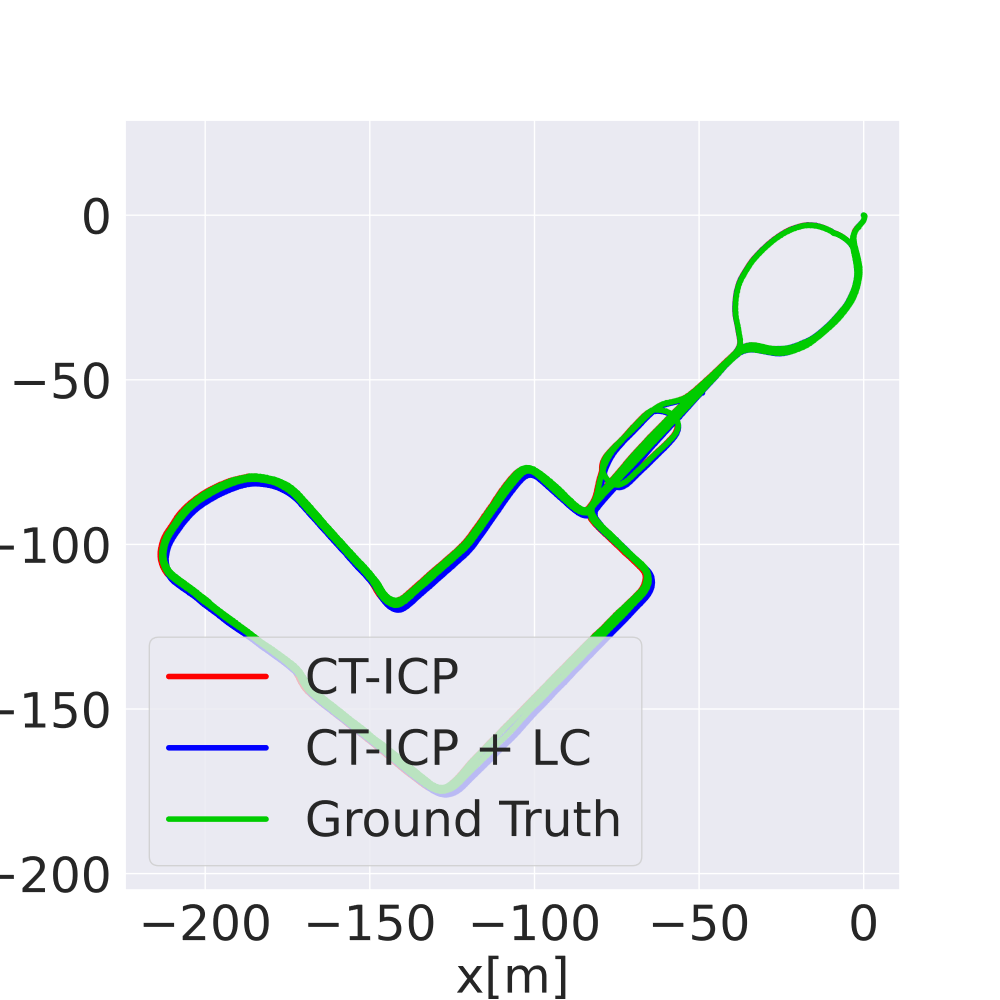}
\includegraphics[width=0.19\linewidth]{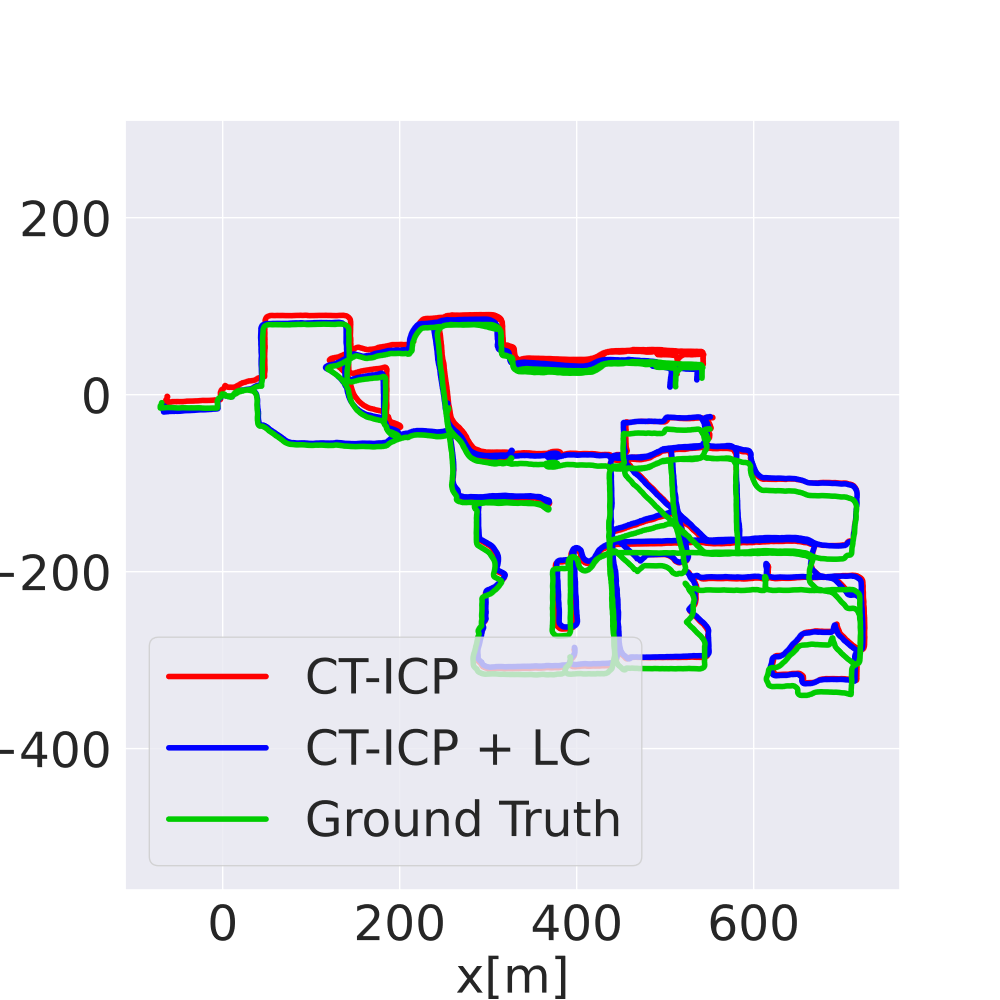}
\caption{\customfigcaption{Trajectories estimated with CT-ICP odometry, and our Loop Closure correction (CT-ICP+LC) for the sequences~\texttt{00} of KITTI-raw (4541 scans),~\texttt{06} of KITTI-360 (9698 scans), \texttt{Town01} of KITTI-CARLA (5000 scans), \texttt{01\_short\_experiment} of NCD (15301 scans) and \texttt{2012-01-08} of NCLT (42764 scans).}} 
\label{fig:trajectories}
\vspace{-20pt}
\end{figure*}
Our CT-ICP odometry is parameterized for the current scan by two poses: a pose for the beginning of the scan $\Omega_b^n=(R^n_b, t^n_b)$ ($R$ for rotation, $t$ for translation, and $b$ for beginning) and a pose for the end of the scan $\Omega_e^n=(R^n_e, t^n_e)$ ($e$ for end). 
To simplify the notations, in the following, we omit the $n$ referring to the poses of the current scan. 
For each sensor measurement captured at a time $\tau\in [\tau_b,\tau_e]$ between the first $\tau_b$ and last $\tau_e$ timestamp of the scan, the pose of the sensor is estimated by interpolating between the two poses of the scan.  
These poses transform a point from the LiDAR frame $L(\tau)$ to the world frame $W=L(0)$. 
Unlike other continuous-time trajectory methods, the pose $\Omega_b$ of the new scan does not match the end pose $\Omega_e^{n-1}$ of the previous scan. 
We add a proximity constraint in the optimization to force the two poses to remain close. Our formulation allows our odometry to be more robust to high-frequency motions of the sensor.

For each new scan $\mathbb{S}^n$, we first extract a sample of keypoints from $\mathbb{S}^n$ indexed by $\mathbb{I}^n$: $\{p_i\in \mathbb{S}^n | i\in \mathbb{I}^n\}$ (using a simple grid sampling of points in the scan), that we register into the local map. This map is a dense point cloud $\mathbb{M}^n=\{q_i^W\}$ built from all the previously registered scans, and stored in a sparse voxel grid. Its construction is detailed in section \ref{local_map} below.  
%and the poses $(R_b^i, t_b^i)$ and $(R_e^i, t_e^i) $ of the previous scans ($i$ from $0$ to $n-1$).
Our scan matching then estimates the two optimal poses $\Omega_b^*$ and $\Omega_e^*$, thus handling the distortion of the scan during the optimization, and proceeds to transform the points to the world frame before adding them to the local map. 

These optimal poses are given by solving the following problem with parameters $\mathbf{X}=(\,\boldsymbol{\Omega_b},\boldsymbol{\Omega_e}\,)\in SE(3)^2$ in bold:
\begin{equation}
\small
\argmin_{\mathbf{X}\in SE(3)^2}\; F_{\text{ICP}}(\,\mathbf{X}\,) + \beta_l \, C_{\text{loc}}(\mathbf{X}) + \beta_v \, C_{\text{vel}}(\mathbf{X})
\label{eq:ct_icp}
\end{equation}
where $F_{\text{ICP}}$ is the scan-to-map continuous-time ICP:
\begin{align}
\small
&F_{\text{ICP}}(\,\mathbf{X}\,) = \frac{1}{\vert \mathbb{I}^n \vert} \sum_{i\in\mathbb{I}^n} \rho (r_i^2[\mathbf{X}])
\end{align}
And, for each $i \in \mathbb{I}^n$, 
\begin{align}
r_i[&\mathbf{X}] = a_{i} \;(\,p_i^W[\mathbf{X}] - q_i^W \,)\cdot n_i \\
p_i^W[&\mathbf{X}]= R^{\alpha_i}[\mathbf{X}] *  p_i^L + t^{\alpha_i}[\mathbf{X}]\\
R^{\alpha_i}[&\mathbf{X}] = \text{slerp}(\,\mathbf{R_b},\, \mathbf{R_e},\, \alpha_i\,) \\
t^{\alpha_i}[&\mathbf{X}] = (1-\alpha_i)\, \mathbf{t_b} + \alpha_i\, \mathbf{t_e}
\end{align}

$\rho(s)$ is a robust loss function to minimize the influence of outliers, and $r_i$ are the residuals of the point-to-plane distance between a sample point $p_i^W$ and its closest neighbor in the map (as classically formulated by the ICP). 
$p_i^W[\mathbf{X}]$ is the point $p_i$ expressed in the world frame, $n_i$ is the normal of $p_i^W$'s neighborhood in the local map, and $p_i^L$ is the sensor measurement (in the LiDAR frame). 
$\boldsymbol{\Omega}^{\alpha_i}[\mathbf{X}]=(\mathbf{R^{\alpha_i}}, \mathbf{t^{\alpha_i}})\in SE(3)$ is the transformation from the LiDAR frame at time $\tau_i$, $L(\tau_i)$ to the world $W$. 
It is estimated by an interpolation between $\boldsymbol\Omega_b$ and $\boldsymbol\Omega_e$ by defining $\alpha_i=(\tau_i-\tau_b)/(\tau_e-\tau_b)$. 
For rotation interpolation, we use the standard spherical linear interpolation (slerp).

We also introduce weights to favor planar neighborhoods: $a_{i} = a_{2D} = (\sigma_2 - \sigma_3)/\sigma_1$ as defined by~\cite{deschaud2018imlsslam} (where $\sigma_i$ are the square roots of the eigen values of the neighborhood's covariance), is the planarity of the neighborhood of $p_i^W$. Note that unlike~\cite{deschaud2018imlsslam} and most LiDAR odometry methods, we calculate the normals $n_i$ and planarity weights $a_i$ using the dense point cloud of the local map instead of the current scan, leading to much richer neighborhoods. This computation is done at each iteration of the ICP and for each sample point $p_i$, as the refinement on their position leads to more precise neighborhoods.

%We also  
%$n_i$ is the estimated normal at the point $p_i^W[\mathbf{X}]$,
%$p_i^L$ the coordinates of the sample $i$ in the LiDAR frame $L$,
%$(R(\alpha_i), t(\alpha_i))$ the transformation from the LiDAR frame $L$ to the world frame $W$ considering the timestamp $\tau_i$ of the point $p_i$ with $\alpha_i=\frac{\tau_e-\tau_i}{\tau_e-\tau_b}$ with $\tau_e$ and $\tau_b$ the timestamps of the first and last points of the current scan,
%$q_i^W$ the point closest to $p_i$ in the local map (in world frame $W$),
%$\text{slerp}$ is the spherical linear interpolation for rotations.

Additionally in equation \ref{eq:ct_icp}, we have introduced the two constraints $C_{loc}$ (location consistency constraint), and $C_{vel}$ (constant velocity constraint), with respective weights $\beta_l$ and $\beta_v$, which is defined as follows:
\begin{align}
\small
C_{loc}(\mathbf{t_b}) &= || \mathbf{t_b} - t_e^{n-1}||^2 \\
C_{vel}(\mathbf{t_b}, \mathbf{t_e}) &= || (\mathbf{t_e} - \mathbf{t_b}) - (t_e^{n-1} - t_b^{n-1}) ||^2 
\end{align}

$C_{loc}$ forces the beginning and end location of the sensor to be consistent (limiting the discontinuity), and $C_{vel}$ limits too rapid accelerations. These constraints are sufficient to force the elasticity of CT-ICP to be consistent with the motion model defined implicitly by the weights $\beta_l$ and $\beta_v$ (taken at 0.001 for all experiments).

CT-ICP performs iterations until a threshold on the norm of parameter's step is met (typically 0.1 cm on translation and 0.01° in rotation) or a number of iterations is reached (5 to ensure real time). By default, CT-ICP runs on a single thread, but parallel processing (for the construction of the neighborhoods, or while solving the linear system) can be leveraged for more challenging scenarios requiring more iterations and sample points.

\subsection{Local map and robust profile} \label{local_map}

%For each scan, before our alignment we sample points (at a finer resolution than the keypoints sampling size), this allows us to limit the number of points belonging to a same scan in the map.

As a local map, we use a point cloud from previous scans (like IMLS-SLAM~\cite{deschaud2018imlsslam}), but in contrast, the points in the world frame ($W$) are stored in a sparse data structure of voxels for faster neighborhood access than kd-trees (constant time access instead of logarithmic).  
The voxel size of the map controls the radius of the neighborhood search, and the level of detail of the stored point cloud. 
For driving scenarios, it is set to 1.0\,m and 0.80\,m for high-frequency motion scenarios, providing satisfying balance between these two aspects.

The voxel size defining the grid of the map is important because it defines the neighborhood search radius, as well as the level of detail of our local map. 

Each voxel stores up to 20 points, such that no two points are closer than 10\,cm from each other, to limit the redundancy due to the density of measurements along the scan lines. 
Once a voxel is full, no more points will be inserted into it.

To construct the neighborhood of a point $p_i^W$ (needed for the computation of $n_i$ and $a_i$), we select the $k=20$ nearest neighbors among all points in the map in the 27 neighboring voxels from the current point.
%, though indoor scenarios would typically require a smaller voxel size.
%Indeed, each voxel has a maximum number of points stored, and no to points can be closer to $d=10cm$ (for outdoor scenarios)
%For all our experiments conducted outdoors we set the voxel size to $1.0m$.
% as access time for a voxel stored by a hash map is constant regardless of the size of the map. 
%No information other than the 3D coordinates of points in the world coordinate system ($W$) is stored in the map. 
%The other characteristics used in the ICP (normal $n_i$, planarity $a_{2D}$) are computed on-the-fly.
%The parameters of the local map are:
%\begin{itemize}
%\item \textit{size\_voxel\_map}: size of a voxel storing the points of the local map and used for the neighborhood calculation during CT-ICP.
%\item \textit{max\_num\_points\_in\_voxel}: maximum number of points in a voxel (to limit the number of points of the local map).
%\item \textit{min\_distance\_points\_in\_voxel}: minimum distance between two points in the same voxel (to better distribute the points of the local map).
%\item \textit{number\_neighbors}: the number of points taken for the definition of neighborhood (in the computation of normal $n_i$ and planarity $a_{2D}$).
%\end{itemize}
After running CT-ICP for the current scan $n$, the points are added to the local map. %Points in voxels that are fully occupied are dropped.
Note that in contrast with pyLiDAR's F2M, the local map is not based on a sliding window of last scans, but voxels are removed from the map based only on their distance to the center of the last inserted scan.

Odometries with these types of maps are highly sensitive to bad registration, and cannot recover from map pollutions from bad scan insertions. 
This is particularly problematic in datasets with rapid orientation changes.
For these types of datasets, we introduce a robust profile that detects hard cases (fast orientation changes) and registration failures (location inconsistencies or large number of new keypoints falling in empty voxels) and attemps a new registration of the current scan with a more conservative set of parameters (most notably a larger number of sampled keypoints and a larger neighborhood search); for important modifications in the orientation ($\ge 5$\,$^{\circ}$), we do not insert the new scan in the map, which has a higher probability of being misaligned. 
The gain in robustness is paid by increased runtime.

%%%%%%%%%%%%%%%%%%%%%%%%%%%%%%%%%%%%%%%%%%%%%%%%%%%%%%%%%%%%%%%%%%%%%%%%%%%%%%%%
\section{LOOP CLOSURE AND BACK-END}
% main author Pierre
%We present in this section a new purely LiDAR loop closure scheme which is robust to orientation changes between new scans and previously explored environments (even with large drift), as section~\ref{loop_closure_exp} will show with experimentations realized on multiple datasets. 
 
\begin{figure}[t]
\includegraphics[width=\linewidth]{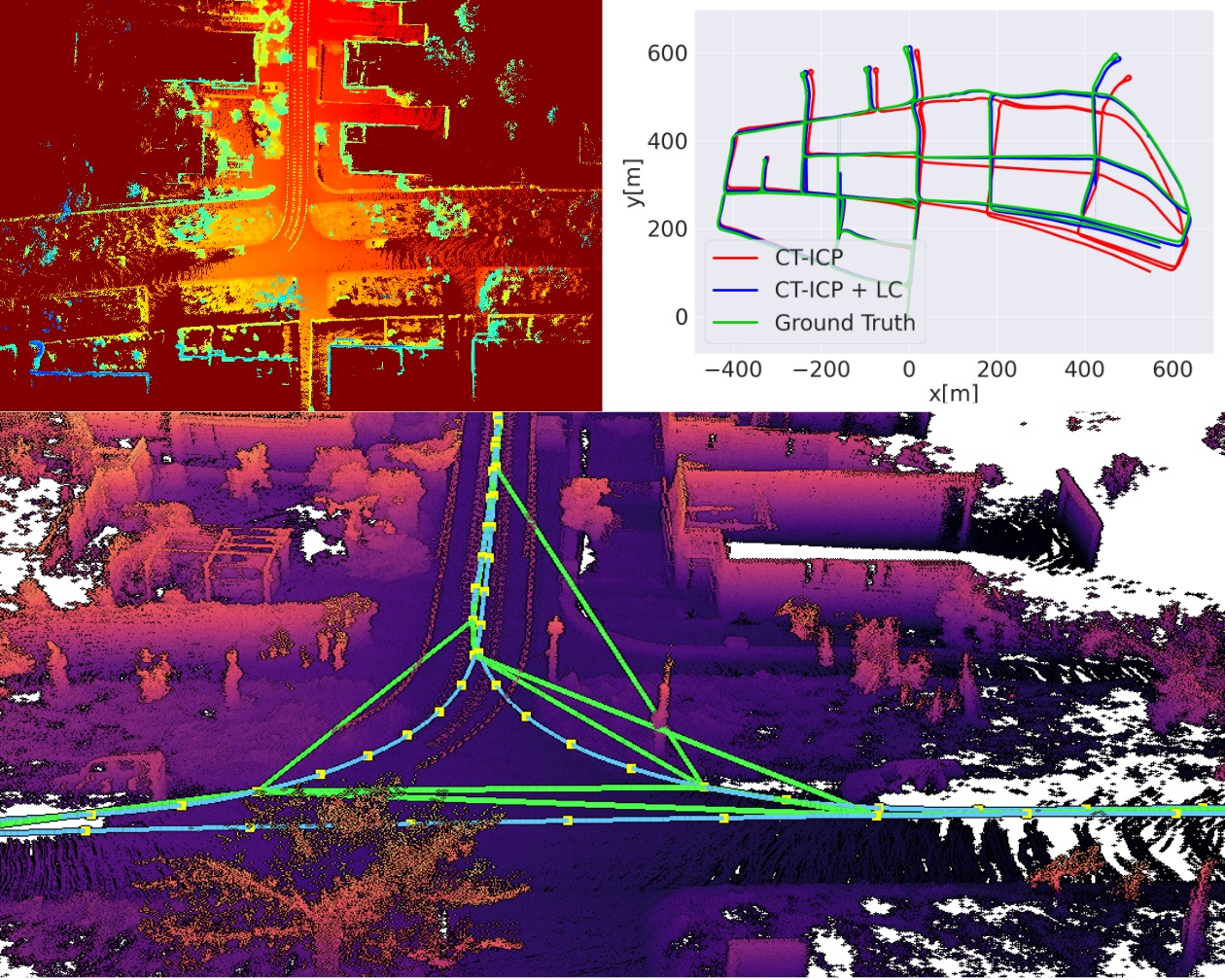}
\caption{\customfigcaption{Qualitative results of loop closure on the sequence~\texttt{00} of KITTI-360 (11501 scans). The top left is an elevation image built by projecting the local map. The top right shows both the CT-ICP odometry's trajectory and the one corrected using the computed Loop Closure constraints (CT-ICP+LC). The bottom shows the different loop closure constraints (green) found for the same turn as the local map at the top left.}}
\label{qualitative_loop_closure}
\vspace{-10pt}
\end{figure}

Our loop closure algorithm maintains in its memory a window of the last scans registered by the odometry. When the window reaches a size of $N_{map}$ scans, the points are aggregated into a point cloud that is placed in the coordinate frame at the center of the window. 

Every point of this map is then inserted into a 2D elevation grid, keeping for each pixel the point at a maximum elevation. From this 2D grid, an elevation image is obtained by clipping the $z$ coordinate of each pixel between $z_{min}$ and $z_{max}$. Rotation invariant 2D features \cite{akaze} are then extracted and saved in memory with the elevation grid. All scans except the last $N_{overlap}$ are dropped from the window. 
 
Every time a new elevation image is built (every $N_{map}-N_{overlap}$ scans), it is matched against the elevation images saved in its memory. 
A 2D rigid transformation between the two features sets is robustly fit with a RANSAC, and a threshold on the number of inliers is used to validate the correspondences. 
When a match is validated, an ICP refinement of the initial 2D transform (using Open3D's ICP~\cite{Zhou2018}) is performed on the elevation grids' point clouds, producing an accurate 6-DoF loop closure constraint. 
To reduce the number of candidates, an initial filter selects the $n_{candidates}=10$ candidates closest to the current grid. 
%Note that this parameter should be adapted depending on the expected drift, which was low enough using our proposed odometry for all our datasets. 

As a back-end, our SLAM uses a standard pose graph (PG) implemented with g2o~\cite{g2o} much like in~\cite{mendes2016icp}. 
The PG regularly adds new poses when new odometry constraints are added with the respective odometry constraints, but the trajectory is only globally optimized when a new loop constraint is detected, in which case the trajectory of the loop closure module is also updated. Fig.~\ref{qualitative_loop_closure} summarizes our loop closure procedure.

Currently, our procedure requires the sensor's motion to be mostly planar on the ground, and for the extrinsic calibration to align the $z$ axis with the normal of the ground plane. This limits the supported sensor setups, as well as the authorized motion. Note that this can be addressed if the gravity vector or the local ground plane are known, in which case the elevation image can be correctly projected. However, we show in section \ref{experiments} that for outdoor scenarios with a sensor mostly pointing up, our loop closure detection succeeds in detecting many loops, allowing our back-end to successfully correct the drift errors from the odometry.

%%%%%%%%%%%%%%%%%%%%%%%%%%%%%%%%%%%%%%%%%%%%%%%%%%%%%%%%%%%%%%%%%%%%%%%%%%%%%%%%
\section{EXPERIMENTS} \label{experiments}
% main author Pierre

We demonstrate the efficiency and versatility of our method by conducting experiments on a large variety of datasets: \KITTI, \KITTI-raw, \newKITTI, \KITTICarla, \ParisLuco, Newer College Dataset (\NCD), and \NCLT. Our method requires only the geometry (\texttt{xyz} fields) and the timestamps for each point to elastically distort the scans (they are estimated linearly based on azimuth angles for \KITTI-raw and \newKITTI).

\subsection{Datasets}

\subsubsection{Driving scenarios}

KITTI~\cite{geiger2012kitti} proposes 11 sequences of LiDAR scans acquired from a Velodyne HDL64 mounted on a car and the corresponding ground truth poses from GPS/IMU measurements. 
The odometry benchmark proposes motion-corrected scans (using the GPS/IMU trajectory to distort the scans): we call this KITTI-corrected. 
Most odometry methods~\cite{zhang2017loam, behley2018suma, deschaud2018imlsslam, pan2021mulls} compare their results on the motion-corrected version of KITTI. 
However, this does not allow for testing their performance with raw LiDAR data. 
All odometry sequences (except sequence \texttt{03}) are available with uncorrected raw scans. 
We call this KITTI-raw. 
The same acquisition setup was later used to generate KITTI-360~\cite{Xie2016CVPR}, which contains eight sequences that are much longer than KITTI's (ranging from 3000 to 15000 scans by sequence) in the same environment. 
These scans are not motion-corrected, and the timestamps are estimated as described above.
%, is acquired using the same acquisition system, in the same city. It proposes 7 sequences, much longer than KITTI's (ranging from 3000 to 15000 scans by sequence), and often containing multiple loops. Precise localization is obtained by large-scale optimization taking GPS/IMU measurements, laser scans and multi-view images but is available only at some timestamps, from which we estimate ground truth poses for every scan by interpolation. The LiDAR scans of KITTI-360 are not motion-corrected, and timestamps are estimated as mentioned above.
For KITTI-corrected, KITTI-raw, and KITTI-360, we applied for all scans an intrinsic angle correction of 0.205\,$^{\circ}$, as mentioned in~\cite{deschaud2018imlsslam}.

KITTI-CARLA~\cite{deschaud2021kitticarla} simulates a similar sensor suite as KITTI in a synthetic environment generated using the CARLA simulator \cite{dosovitskiy2017carla}. 
The dataset consists of seven sequences of 5000 scans using a simulated 64-channel LiDAR sensor. The relative motion of the vehicule during the acquisition of the scan is simulated, and each scan provides precise ground truth and timestamps.

ParisLuco is one sequence of 4\,km acquisition (12751 scans) in the center of Paris using our own vehicle with a Velodyne HDL32 sensor in a vertical position. 
We generated the ground truth with post-processing of GPS/IMU measurements (only translations are available).

%\textit{UrbanLoco~\cite{urban_loco}} proposes two sensor acquisition setup acquiring data in San Francisco (7 sequences using a 32 channel RobotSense LiDAR) and Hong-Kong (3 sequences with a Velodyne HDL32 LiDAR). The dataset does not provide the full 6DoF poses in the coordinate frame of the LiDAR, but the measurements provided by the GNSS/INS novatel navigation system allow an accurate estimation of the \texttt{X,Y} components of the trajectory, expressed in the ENU local frame. Timestamps for each point (which are not fournished) are estimated linearly by retrieving the Velodyne/RobotSense packet ids of each point. Finally, the scans are cut similarly to KITTI (in order to have the begining and end of a scan behind the car).

\subsubsection{High-frequency motion scenarios}

%Additionnally, we tested our method on two challenging non-driving datasets. 

Cars accelerate slowly with respect to the speed of acquisition of the sensor; thus, a constant velocity model is valid. 
Much more challenging are scenarios were the sensor quickly changes orientations. 
This is the case for NCLT~\cite{ncarlevaris-2015a}, a dataset containing 27 long sequences ($\ge20000$ scans per sequence) of scans acquired at Michigan University using a Velodyne HDL32 mounted on a two-wheeled Segway. 
This dataset is particularly challenging, because the vehicule introduces abrupt rotations to the LiDAR's own rotation axis, which is problematic for classical ICP-based odometry methods, as shown in~\cite{dellenbach2021}

NCD~\cite{ncd}, the Newer College Dataset, contains two sequences ($\sim$15000 and $\sim$26000 scans) of a handheld Ouster 64-channel LiDAR mounted on a stick, that is carried accross the Oxford campus.

\subsection{Odometry Experiments}

\begin{table*}[t]
\caption{\customtabcaption{Relative Translation Error (\textit{RTE}) [\%] with the \textbf{Driving profile} on KITTI-corrected, KITTI-raw, KITTI-360, KITTI-CARLA, and ParisLuco and with the \textbf{High-Frequency Motion profile} on the NCD and NCLT datasets. AVG is the \textit{RTE} averaged over all segments of all sequences, $\Delta T$ the average running time per scan. KITTI-corrected is the only dataset whose scans have been motion-corrected $(\star)$; all other datasets have uncorrected raw point clouds as scans.}}

\begin{center}
\scriptsize
\begin{tabular}{p{2.3cm}|*{22}{p{0.1cm}}p{0.35cm}p{0.35cm}}
\multicolumn{22}{l}{\textbf{DRIVING PROFILE}}\\
\toprule 
\toprule
\textbf{KITTI-corrected}$^{\star}$ & \multicolumn{2}{c|}{\texttt{00}} & \multicolumn{2}{c|}{\texttt{01}} & \multicolumn{2}{c|}{\texttt{02}} & \multicolumn{2}{c|}{\texttt{03}} & \multicolumn{2}{c|}{\texttt{04}} & \multicolumn{2}{c|}{\texttt{05}} & \multicolumn{2}{c|}{\texttt{06}} & \multicolumn{2}{c|}{\texttt{07}} & \multicolumn{2}{c|}{\texttt{08}} & \multicolumn{2}{c|}{\texttt{09}} & \multicolumn{2}{c|}{\texttt{10}}&  \multicolumn{1}{c}{\col{\textbf{AVG}}} & \multicolumn{1}{c}{$\Delta T$} \\ 
\midrule
IMLS-SLAM~\cite{deschaud2018imlsslam} & \multicolumn{2}{c}{0.50} & \multicolumn{2}{c}{0.82} & \multicolumn{2}{c}{0.53} & \multicolumn{2}{c}{0.68} & \multicolumn{2}{c}{\textbf{0.33}} & \multicolumn{2}{c}{0.32} & \multicolumn{2}{c}{0.33} & \multicolumn{2}{c}{0.33} & \multicolumn{2}{c}{0.80} & \multicolumn{2}{c}{0.55} & \multicolumn{2}{c}{0.53} & \multicolumn{1}{c}{\col{0.55}} & \multicolumn{1}{c}{1250\,ms} \\
MULLS~\cite{pan2021mulls} & \multicolumn{2}{c}{0.56} & \multicolumn{2}{c}{\textbf{0.64}} & \multicolumn{2}{c}{0.55} & \multicolumn{2}{c}{0.71} & \multicolumn{2}{c}{0.41} & \multicolumn{2}{c}{0.30} & \multicolumn{2}{c}{0.30} & \multicolumn{2}{c}{0.38} & \multicolumn{2}{c}{\textbf{0.78}} & \multicolumn{2}{c}{0.48} & \multicolumn{2}{c}{0.59} & \multicolumn{1}{c}{\col{0.55}} & \multicolumn{1}{c}{80\,ms} \\
pyLiDAR F2M~\cite{dellenbach2021} & \multicolumn{2}{c}{0.51} & \multicolumn{2}{c}{0.79}& \multicolumn{2}{c}{\textbf{0.51}}& \multicolumn{2}{c}{\textbf{0.64}}& \multicolumn{2}{c}{0.36}& \multicolumn{2}{c}{0.29}& \multicolumn{2}{c}{0.29} & \multicolumn{2}{c}{0.32}& \multicolumn{2}{c}{\textbf{0.78}} & \multicolumn{2}{c}{\textbf{0.46}}& \multicolumn{2}{c}{0.57} & \multicolumn{1}{c}{\col{\underline{\textbf{0.53}}}} &  \multicolumn{1}{c}{175\,ms}\\
CT-ICP (ours) & \multicolumn{2}{c}{\textbf{0.49}} & \multicolumn{2}{c}{0.76} & \multicolumn{2}{c}{0.52} & \multicolumn{2}{c}{0.72} & \multicolumn{2}{c}{0.39} & \multicolumn{2}{c}{\textbf{0.25}} & \multicolumn{2}{c}{\textbf{0.27}} & \multicolumn{2}{c}{\textbf{0.31}} & \multicolumn{2}{c}{0.81} & \multicolumn{2}{c}{0.49} & \multicolumn{2}{c}{\textbf{0.48}} & \multicolumn{1}{c}{\col{\underline{\textbf{0.53}}}} & \multicolumn{1}{c}{60\,ms} \\
\midrule
%\noalign{\vskip 1mm}

\textbf{KITTI-raw} & \multicolumn{2}{c|}{\texttt{00}} & \multicolumn{2}{c|}{\texttt{01}} & \multicolumn{2}{c|}{\texttt{02}} & \multicolumn{2}{c|}{\texttt{X}} & \multicolumn{2}{c|}{\texttt{04}} & \multicolumn{2}{c|}{\texttt{05}} & \multicolumn{2}{c|}{\texttt{06}} & \multicolumn{2}{c|}{\texttt{07}} & \multicolumn{2}{c|}{\texttt{08}} & \multicolumn{2}{c|}{\texttt{09}} & \multicolumn{2}{c|}{\texttt{10}} & \multicolumn{1}{c}{\col{\textbf{AVG}}} & \multicolumn{1}{c}{$\Delta T$} \\ 
\midrule
IMLS-SLAM~\cite{deschaud2018imlsslam} & \multicolumn{2}{c}{0.79} & \multicolumn{2}{c}{0.86} & \multicolumn{2}{c}{0.76} & \multicolumn{2}{c}{} & \multicolumn{2}{c}{0.51} & \multicolumn{2}{c}{0.48} & \multicolumn{2}{c}{0.62} & \multicolumn{2}{c}{0.67} & \multicolumn{2}{c}{0.97} & \multicolumn{2}{c}{0.70} & \multicolumn{2}{c}{0.75} & \multicolumn{1}{c}{\col{0.71}} & \multicolumn{1}{c}{1070\,ms} \\
MULLS~\cite{pan2021mulls} & \multicolumn{2}{c}{1.43} & \multicolumn{2}{c}{3.12} & \multicolumn{2}{c}{1.01} & \multicolumn{2}{c}{} &\multicolumn{2}{c}{0.57} & \multicolumn{2}{c}{1.93} & \multicolumn{2}{c}{1.60} & \multicolumn{2}{c}{0.69} & \multicolumn{2}{c}{1.28} & \multicolumn{2}{c}{1.49} & \multicolumn{2}{c}{0.71} & \multicolumn{1}{c}{\col{1.41}} & \multicolumn{1}{c}{80\,ms} \\
pyLiDAR F2M~\cite{dellenbach2021} & \multicolumn{2}{c}{2.20} &  \multicolumn{2}{c}{0.98} & \multicolumn{2}{c}{1.55} & \multicolumn{2}{c}{} &  \multicolumn{2}{c}{0.45} & \multicolumn{2}{c}{1.46} & \multicolumn{2}{c}{0.69}  & \multicolumn{2}{c}{1.72} & \multicolumn{2}{c}{1.60} &  \multicolumn{2}{c}{1.28} &  \multicolumn{2}{c}{1.18} & \multicolumn{1}{c}{\col{1.61}} & \multicolumn{1}{c}{530\,ms}\\
CT-ICP (ours) & \multicolumn{2}{c}{\textbf{0.51}} & \multicolumn{2}{c}{\textbf{0.81}} & \multicolumn{2}{c}{\textbf{0.55}} & \multicolumn{2}{c}{} & \multicolumn{2}{c}{\textbf{0.43}} & \multicolumn{2}{c}{\textbf{0.27}} & \multicolumn{2}{c}{\textbf{0.28}} & \multicolumn{2}{c}{\textbf{0.35}} & \multicolumn{2}{c}{\textbf{0.80}} & \multicolumn{2}{c}{\textbf{0.47}} & \multicolumn{2}{c}{\textbf{0.49}} & \multicolumn{1}{c}{\col{\underline{\textbf{0.55}}}} & \multicolumn{1}{c}{65\,ms} \\
\midrule
%\noalign{\vskip 1mm}

\textbf{KITTI-360} & \multicolumn{2}{c|}{\texttt{00}} & \multicolumn{2}{c|}{\texttt{02}} & \multicolumn{2}{c|}{\texttt{03}} & \multicolumn{2}{c|}{\texttt{04}} & \multicolumn{2}{c|}{\texttt{05}} & \multicolumn{2}{c|}{\texttt{06}} & \multicolumn{2}{c|}{\texttt{07}} & \multicolumn{2}{c|}{\texttt{09}} & \multicolumn{2}{c|}{\texttt{10}} & & & & & \multicolumn{1}{|c}{\col{\textbf{AVG}}} & \multicolumn{1}{c}{$\Delta T$} \\ 
\midrule
IMLS-SLAM~\cite{deschaud2018imlsslam} & \multicolumn{2}{c}{0.65} & \multicolumn{2}{c}{0.63} & \multicolumn{2}{c}{0.64} & \multicolumn{2}{c}{0.89} & \multicolumn{2}{c}{0.63} & \multicolumn{2}{c}{0.70} & \multicolumn{2}{c}{0.54} & \multicolumn{2}{c}{0.67} & \multicolumn{2}{c}{0.79} & \multicolumn{4}{c}{} &\multicolumn{1}{c}{\col{0.68}} & \multicolumn{1}{c}{1060\,ms}\\
MULLS~\cite{pan2021mulls} & \multicolumn{2}{c}{1.60} & \multicolumn{2}{c}{1.29} & \multicolumn{2}{c}{0.87} & \multicolumn{2}{c}{1.64} & \multicolumn{2}{c}{1.27} & \multicolumn{2}{c}{1.48} & \multicolumn{2}{c}{6.25} & \multicolumn{2}{c}{1.28} & \multicolumn{2}{c}{0.88} & \multicolumn{4}{c}{} &\multicolumn{1}{c}{\col{1.55}} & \multicolumn{1}{c}{90\,ms}\\
pyLiDAR F2M~\cite{dellenbach2021} & \multicolumn{2}{c}{1.79} & \multicolumn{2}{c}{1.25} & \multicolumn{2}{c}{0.90} & \multicolumn{2}{c}{1.60} & \multicolumn{2}{c}{1.26}& \multicolumn{2}{c}{1.38} & \multicolumn{2}{c}{0.69} & \multicolumn{2}{c}{1.72} & \multicolumn{2}{c}{1.39} & \multicolumn{4}{c}{} &\multicolumn{1}{c}{\col{1.46}} & \multicolumn{1}{c}{475\,ms}\\
CT-ICP (ours)& \multicolumn{2}{c}{\textbf{0.41}} & \multicolumn{2}{c}{\textbf{0.38}} & \multicolumn{2}{c}{\textbf{0.34}} & \multicolumn{2}{c}{\textbf{0.65}} & \multicolumn{2}{c}{\textbf{0.39}} & \multicolumn{2}{c}{\textbf{0.42}} & \multicolumn{2}{c}{\textbf{0.34}} & \multicolumn{2}{c}{\textbf{0.45}} & \multicolumn{2}{c}{\textbf{0.69}} & & & & & \multicolumn{1}{c}{\col{\underline{\textbf{0.45}}}} & \multicolumn{1}{c}{70\,ms} \\
\midrule
%\noalign{\vskip 1mm}

\textbf{KITTI-CARLA} & \multicolumn{3}{c|}{\texttt{Town01}} & \multicolumn{3}{c|}{\texttt{Town02}} & \multicolumn{3}{c|}{\texttt{Town03}} & \multicolumn{3}{c|}{\texttt{Town04}} & \multicolumn{3}{c|}{\texttt{Town05}} & \multicolumn{3}{c|}{\texttt{Town06}} & \multicolumn{3}{c|}{\texttt{Town07}} & \multicolumn{1}{c}{}& \multicolumn{1}{|c}{\col{\textbf{AVG}}} & \multicolumn{1}{c}{$\Delta T$} \\ 
\midrule
IMLS-SLAM~\cite{deschaud2018imlsslam} & \multicolumn{3}{c}{\textbf{0.03}} & \multicolumn{3}{c}{0.05} & \multicolumn{3}{c}{0.16} & \multicolumn{3}{c}{0.20} & \multicolumn{3}{c}{0.06} & \multicolumn{3}{c}{4.90} & \multicolumn{3}{c}{\textbf{0.25}} & \multicolumn{1}{c}{}& \multicolumn{1}{c}{\col{0.81}} & \multicolumn{1}{c}{780\,ms} \\
MULLS~\cite{pan2021mulls} & \multicolumn{3}{c}{1.39} & \multicolumn{3}{c}{0.77} & \multicolumn{3}{c}{0.69} & \multicolumn{3}{c}{1.24} & \multicolumn{3}{c}{1.13} & \multicolumn{3}{c}{0.98} & \multicolumn{3}{c}{0.97} & \multicolumn{1}{c}{} & \multicolumn{1}{c}{\col{1.04}} & \multicolumn{1}{c}{70\,ms} \\
pyLiDAR F2M~\cite{dellenbach2021} & \multicolumn{3}{c}{16.25} &  \multicolumn{3}{c}{7.92} & \multicolumn{3}{c}{37.16} & \multicolumn{3}{c}{10.06} & \multicolumn{3}{c}{9.11} & \multicolumn{3}{c}{73.29} &  \multicolumn{3}{c}{2.69} & & \multicolumn{1}{c}{\col{23.84}} & \multicolumn{1}{c}{530\,ms}\\
CT-ICP (ours) & \multicolumn{3}{c}{\textbf{0.03}} & \multicolumn{3}{c}{\textbf{0.04}} & \multicolumn{3}{c}{\textbf{0.03}} & \multicolumn{3}{c}{\textbf{0.03}} & \multicolumn{3}{c}{\textbf{0.02}} & \multicolumn{3}{c}{\textbf{0.04}} & \multicolumn{3}{c}{0.41} & \multicolumn{1}{c}{}& \multicolumn{1}{c}{\col{\underline{\textbf{0.09}}}} & \multicolumn{1}{c}{65\,ms} \\
\midrule
%\noalign{\vskip 1mm}

\textbf{ParisLuco} & \multicolumn{10}{c|}{\texttt{LuxembourgGarden}} & \multicolumn{10}{c}{} & & &    \multicolumn{1}{|c}{\col{\textbf{AVG}}} & \multicolumn{1}{c}{$\Delta T$} \\ 
\midrule
IMLS-SLAM~\cite{deschaud2018imlsslam} & \multicolumn{10}{c}{1.26} & \multicolumn{10}{c}{} & & &    \multicolumn{1}{c}{\col{1.26}} & \multicolumn{1}{c}{770\,ms} \\
MULLS~\cite{pan2021mulls} & \multicolumn{10}{c}{3.03} & \multicolumn{10}{c}{} & & &    \multicolumn{1}{c}{\col{3.03}} & \multicolumn{1}{c}{55\,ms} \\
pyLiDAR F2M~\cite{dellenbach2021}  & \multicolumn{10}{c}{4.90} & \multicolumn{10}{c}{} & & &    \multicolumn{1}{c}{\col{4.90}} & \multicolumn{1}{c}{355\,ms} \\
CT-ICP (ours)  & \multicolumn{10}{c}{\textbf{1.11}} & \multicolumn{10}{c}{} & & &    \multicolumn{1}{c}{\col{\underline{\textbf{1.11}}}} & \multicolumn{1}{c}{80\,ms} \\
 \bottomrule
\noalign{\vskip 1mm}
\multicolumn{14}{l}{}\\

\multicolumn{14}{l}{\textbf{HIGH-FREQUENCY MOTION PROFILE}}\\
\toprule
\toprule
\textbf{NCD} & \multicolumn{10}{c|}{\texttt{01\_short\_experiment}} & \multicolumn{10}{c|}{\texttt{02\_long\_experiment}} & & &   \multicolumn{1}{|c}{\col{\textbf{AVG}}} & \multicolumn{1}{c}{$\Delta T$} \\ 
\midrule
pyLiDAR F2M~\cite{dellenbach2021}  & \multicolumn{10}{c}{1.37}& \multicolumn{10}{c}{2.63} & & & \multicolumn{1}{c}{\col{2.20}} & \multicolumn{1}{c}{1065\,ms}\\
CT-ICP (ours) & \multicolumn{10}{c}{\textbf{0.48}} & \multicolumn{10}{c}{\textbf{0.58}}& & & \multicolumn{1}{c}{\col{\underline{\textbf{0.55}}}} & \multicolumn{1}{c}{430\,ms}\\
\midrule
\noalign{\vskip 1mm}

\textbf{NCLT} & \multicolumn{10}{c|}{\texttt{2012-01-08}} & \multicolumn{10}{c}{\iffalse 2012-04-29\fi} & & &  \multicolumn{1}{|c}{\col{\textbf{AVG}}} & \multicolumn{1}{c}{$\Delta T$} \\ 
\midrule
pyLiDAR F2M~\cite{dellenbach2021} & \multicolumn{10}{c}{19.80} & \multicolumn{12}{c}{} & \multicolumn{1}{c}{\col{19.80}} & \multicolumn{1}{c}{340\,ms}\\
CT-ICP (ours) & \multicolumn{10}{c}{\textbf{1.17}} & \multicolumn{12}{c}{}& \multicolumn{1}{c}{\col{\underline{\textbf{1.17}}}}& \multicolumn{1}{c}{180\,ms}\\
 \bottomrule
\end{tabular}
\end{center}

\vspace{\tabspace}
\vspace{+10pt}
\label{mte_seq_by_seq}
\end{table*}

%First, we conduct experiments to quantitavely evaluate our odometry on all presented datasets. 

For quantitative evaluations, as in~\cite{behley2018suma, deschaud2018imlsslam, pan2021mulls, dellenbach2021}, we use the KITTI Relative Translation Error (\textit{RTE}), which averages the trajectory drift over segments of lengths ranging from 100\,m to 800\,m. %(normalized by the number of scans of the segment). 
When computing the score on multiple sequences, the average (AVG) is computed over all the segments of all sequences, \iffalse (different from the mean of \textit{RTE} over the sequences) \fi and mirrors KITTI's benchmark evaluation.
%Though it is often coupled with the Relative Rotation Error (\textit{RRE}), in practice the \textit{RTE} is sufficient to discriminate algorithms. For the ParisLuco dataset, as the full 6-DoF pose is not available, the drift is only computed using the translations.

To compare our method, we evaluated three other LiDAR odometries on the datasets presented: MULLS~\cite{pan2021mulls}, IMLS-SLAM~\cite{deschaud2018imlsslam}, and pyLIDAR-SLAM F2M~\cite{dellenbach2021}. For each of these methods, we searched for the set of parameters offering the best results, but once the parameters are found, we kept the same for all the driving datasets. MULLS and IMLS-SLAM are odometries specialized for driving datasets, which is why we only compared our odometry to pyLIDAR-SLAM F2M~\cite{dellenbach2021} on the high-frequency motion datasets NCD and NCLT.\\
%MULLS, IMLS-SLAM and pyLIDAR-SLAM use constant velocity type motion models to distort raw scans.

We can see from Table~\ref{mte_seq_by_seq} that all the methods are very close on the KITTI-corrected dataset. % with the motion-corrected scans. 
However, when we use the uncorrected scans (KITTI-raw and KITTI-360), all the other methods strongly degrade their performance. 
On the other hand, CT-ICP manages to obtain results very close to those of KITTI-corrected. 
The corrected scans prevent CT-ICP from using its elastic formulation, but this shows that our map representation and our scan-to-map algorithm are efficient. 
We also launched our odometry on the KITTI-corrected test sequences (from \texttt{11} to \texttt{21}) and submitted the results online on the KITTI benchmark: we are first among those who put their code online, with a score of 0.59\% \textit{RTE}. 

%However, we encourage to see new online benchmarks appear in the future with uncorrected raw LiDAR scans to better assess pure LiDAR odometries.
It should be noted that in Table~\ref{mte_seq_by_seq}, we do not obtain the same results published by MULLS on the KITTI-corrected dataset because we are using a single set of parameters for all sequences, whereas MULLS had three different ones (urban, highway, and country).

%In table \ref{mte_seq_by_seq}, we present the aggregate \textit{MTE} computed for each dataset, sequence by sequence, for both our solvers (using the same driving profile), as well as for MULLS \cite{pan2021mulls} which is the current state of the art on KITTI's dataset \PIER{Depends on MULLS results}. The table show that MULLS performs slightly better on KITTI corrected dataset, however our method is superior in other driving scenarios, notably when taking into account the distortion of the scan is required. 

%(\PIER{Results on NCLT}). 
%On NCLT, we found that MULLS fails before finishing the sequence despite our best effort to search for appropriate parameters, while our method, with CERES solver performs well, as shown quantitatively and qualitatively (see figure \ref{nclt}). 

For KITTI-CARLA, a synthetic dataset with scans comprising greater velocity variations than the KITTI datasets (due to the simulated dynamics of the vehicle), the difference between CT-ICP and the other odometries is more obvious: 0.09\% \textit{RTE} for CT-ICP and 0.81\% for the best of the three other odometries. 
This shows the interest of our elastic formulation and discontinuity of poses between scans and how it is able to compensate for small errors in the poses of previous scans.

ParisLuco is a dataset with a different sensor (32 fibers instead of 64 for KITTI's) and low inertia because it was acquired in the center of a city. 
CT-ICP can overtake IMLS-SLAM, despite using a much sparser map.

NCD and NCLT are challenging datasets due to the platforms' unstability (a stick and segway), their lengths (26000 scans for sequence \texttt{02\_long\_experiment}, and 46257 scans for sequence \texttt{2012-01-08}), and the variety of the environments (vegetation, interior, roads, outdoor), and the large number of challenging edge cases. 
% Further, these datasets have a large number of edge cases, which are typically difficult to handle for typical LiDAR odometries, including rapid changes in environment (from indoor to outdoor), sharp turns around the rotation axis, and large distortion of the scan caused by the high-frequency discontinuity of the motion. 
Table \ref{mte_seq_by_seq} shows that CT-ICP obtains excellent results on NCD (with comparable error rates to KITTI). Further results on NCLT show the robustness of the method because it is able to properly handle all edge cases, resulting in a much lower \textit{RTE} than pyLiDAR F2M.

To isolate our main contribution, the elastic formulation of CT-ICP, we modified the scan matching of our odometry by distorting the scan with a constant velocity model prior the optimization, and estimated a single pose (while CT-ICP refines the distortion by estimating two poses). We went from 0.55\% \textit{RTE} to 0.79\% for KITTI-raw and from 0.45\% \textit{RTE} to 0.60\% for KITTI-360, showing a significant loss in precision.

\subsection{Loop closure experiments} \label{loop_closure_exp}
\begin{table}
\caption{\customtabcaption{Loop Closure (LC) metrics on one sequence of each dataset. \textit{ATE} = mean Absolute Trajectory Error\,[m] after rigid transform is fitted between the ground truth and estimated trajectory, $N_{loop}$ the number of loops detected, LO = LiDAR Odometry}}
\begin{center}
\scriptsize
\begin{tabular}{l| l| c c c c}
 Dataset & Sequence & $N_{loop}$ & ATE (LO) & ATE (LO+LC)  \\ 
 \toprule
  \midrule
 KITTI-raw & \texttt{00}  & 69 & 6.22 & \textbf{0.66} \\  
 \midrule
 KITTI-360 & \texttt{00}  & 234 & 29.87 & \textbf{1.07}  \\ 
 \midrule
 KITTI-CARLA & \texttt{Town01}  & 78 & \textbf{0.21} & 0.26 \\  
 \midrule
 ParisLuco & \texttt{LuxembourgG.} & 292 & 29.16 & \textbf{9.65} \\  
 \midrule
 NCD & \texttt{01\_short\_exp.} & 59 & \textbf{0.22} & 0.36 \\  
 \midrule 
 NCLT & \texttt{2012-01-08} & 520 & 2.97 & \textbf{2.58} \\
\bottomrule
\end{tabular}
\label{table_loop_closure}
\end{center}
\vspace{-20pt}
\end{table}

For all datasets, we run CT-ICP with the loop closure module (CT-ICP+LC) with $N_{map}=100$ and $N_{overlap}=30$. 
%Our algorithm detected appropriate loops for all sequences in which the vehicule passed by the same location for each dataset. 
For every dataset, we set $z_{min}$ slightly below the ground (using an approximate extrinsic calibration of the sensor) and $z_{max}=z_{min}+10$\,m. Finally, for NCLT, we apply a rigid transform to have the default $z$ axis pointing up (instead of down). 
The computation of each elevation grid and matching of previous ones requires 1.1\,s on average, and the pose graph optimization (only launched when a loop constraint is detected) requires an additional 1.2\,s on average. 
% [Implementation detail glossed over] Our loop closure currently runs on the main thread; however, using separate threads is easily achievable and would lead to real-time complete SLAM. 

We present qualitative results of our loop closure in Fig.~\ref{fig:trajectories} and  Fig.~\ref{qualitative_loop_closure}.
To quantitavely evaluate its quality, the \textit{RTE} is not well suited and tends to deteriorate after loop closure, as noted by~\cite{pan2021mulls}. 
%Having a loop closure which does not deteriorate the local quality of the map is a research topic beyond the scope of this work (notably addressed by~\cite{hatem2014continous}), 
%To focus on evaluating the global trajectory improvements $i.e.$ the correction of the LiDAR odometry's drift coming from the loop closure.
To prove global trajectory improvements, we use the standard Absolute Trajectory Error (\textit{ATE}) to evaluate our method. However, to decouple the quality assessment from orientation errors in the initial poses, 
we first estimate the best rigid transform between the ground truth and the estimated trajectory before computing the \textit{ATE}.
%and apply it to the latter. This leads to a good assessment of the overall quality of the results for one sequence of each dataset are presented in Table~\ref{table_loop_closure}.

Table~\ref{table_loop_closure} shows the results for a sequence of each dataset. First, we note the great improvement of the \textit{ATE} after loop closure for KITTI-raw, KITTI-360, and ParisLuco; more so on KITTI-360 because the sequence is much longer. 
On KITTI-CARLA however, the \textit{ATE} is almost the same before or after loop closure. 
This dataset has a very simple geometry, with large and perfect planes, so the challenge to the scan-matching is principally the motion of the sensor during acquisition. 
CT-ICP manages near-perfect alignments and already leads to 21\,cm of precision on the absolute trajectory estimation; this is visualized in Fig.~\ref{fig:trajectories}, where the ground truth and estimated trajectories are indistinguishable. 
%This is due to the limitation of our back-end, which is designed to correct large trajectory errors, but does so at the price of the local point cloud quality.

%However the relative poses estimated by our loop closure algorithm are less precise as they result from a point-to-point alignment based on elevation grid, they add noise imprecision which slightly deteriorates the metric.
%This however is tied to the limitation of our backend, which is designed to correct large trajectory errors, but is not well suited for preserving the intrinsic quality of the point cloud. 
Finally, we note the relative large number of loop closure constraints detected for each sequence. This number depends on the frequency of the construction of elevation grids, here as controlled by the size of the overlap $N_{overlap}=30$ and the size of the map $N_{map}=100$. 
When arriving at an intersection, multiple loop constraints can be computed, as shown in Fig.~\ref{qualitative_loop_closure}, and with a larger overlap, our method could detect more loop constraints. % (including the constraints between consecutive local maps). 
\section{CONCLUSION}

We presented a new real-time LiDAR-only odometry method that goes beyond the state of the art on seven datasets with different profiles, from driving to high-frequency motion scenarios. 
At the core of our method is the continuous scan matching CT-ICP, which elastically distorts a new scan during the optimization to compensate for the motion during the acquisition.
We publish both the code and datasets to allow for the reproduction of all the results presented above.
%Our LiDAR odometry is robust and precise. 
In future work, we will focus on improving our back-end to extend the proposed continuous formulation beyond the scan matching and to fully leverage our loop closure procedure.

\clearpage

\bibliographystyle{IEEEtran}
\bibliography{IEEEabrv,biblio}

\end{document}